\documentclass[a4paper,twoside]{article}

\usepackage{epsfig}
\usepackage{subcaption}
\usepackage{calc}
\usepackage{amssymb}
\usepackage{amstext}
\usepackage{amsmath}
\usepackage{amsthm}
\usepackage{multicol}
\usepackage{pslatex}
\usepackage{apalike}
\usepackage[bottom]{footmisc}
\usepackage[hyphens]{url}
\usepackage{hyperref}
\hypersetup{breaklinks=true}
\usepackage{censor,caption}
\usepackage{SCITEPRESS}     

\begin{document}

\title{German BERT Model for Legal Named Entity Recognition}

\author{\authorname{Harshil Darji\orcidAuthor{0000-0002-8055-1376}, Jelena Mitrović\orcidAuthor{0000-0003-3220-8749} and Michael Granitzer\orcidAuthor{0000-0003-3566-5507}}
\affiliation{Chair of Data Science, University of Passau, Innstraße 41, 94032 Passau, Germany}
\email{\{harshil.darji, jelena.mitrovic, michael.granitzer\}@uni-passau.de}}


\abstract{The use of BERT, one of the most popular language models, has led to improvements in many Natural Language Processing (NLP) tasks. One such task is Named Entity Recognition (NER) i.e. automatic identification of named entities such as location, person, organization, etc. from a given text. It is also an important base step for many NLP tasks such as information extraction and argumentation mining. Even though there is much research done on NER using BERT and other popular language models, the same is not explored in detail when it comes to Legal NLP or Legal Tech. Legal NLP applies various NLP techniques such as sentence similarity or NER specifically on legal data. There are only a handful of models for NER tasks using BERT language models, however, none of these are aimed at legal documents in German. In this paper, we fine-tune a popular BERT language model trained on German data (German BERT) on a Legal Entity Recognition (LER) dataset. To make sure our model is not overfitting, we performed a stratified 10-fold cross-validation. The results we achieve by fine-tuning German BERT on the LER dataset outperform the BiLSTM-CRF+ model used by the authors of the same LER dataset. Finally, we make the model openly available via HuggingFace.}

\onecolumn \maketitle \normalsize \setcounter{footnote}{0} \vfill

\section{\uppercase{Introduction}}
\label{sec:introduction}

In NLP, NER is the automatic identification of named entities in unstructured data. These named entities are assigned to a set of semantic categories \cite{grishman1996message}, for example, for German Wikipedia and online news, such semantic categories are, \textit{Location} (\textbf{LOC}), \textit{Organization} (\textbf{ORG}), \textit{Person} (\textbf{PER}), and \textit{Other} (\textbf{OTH}) \cite{benikova2014nosta}. However, these named entities are not compatible with the legal domain because the legal domain also contains some domain-specific named entities such as judges, courts, court decisions, etc. A NER model fine-tuned on such domain-specific data improves the efficiency of researchers or employees working on such documents.

In the past couple of decades, there have been many improvements in terms of approaches being used for NER. From standard linear statistical models such as Hidden Markov Model \cite{mayfield2003named,morwal2012named} to CRFs \cite{lafferty2001conditional,finkel2005incorporating,benikova2015c}, RNNs \cite{chowdhury2018multitask,li2020wcp}, and BiLSTMs \cite{huang2015bidirectional,lample2016neural}. However, the introduction of Transformers \cite{vaswani2017attention} gave rise to more efficient tools for NLP, such as BERT \cite{devlin2018bert}, RoBERTa \cite{liu2019roberta}, etc. With these improved language models, there has been a significant improvement in terms of research in NER. Nowadays, many BERT language models take advantage of their underlying transformer approach to produce a specific BERT model fine-tuned for NER tasks in different languages \cite{souza2019portuguese,labusch2019bert,jia2020entity,taher2020beheshti}. There is also research done for BERT in the legal domain that uses BERT for various legal tasks such as topic modeling \cite{silveira2021topic}, legal norm retrieval \cite{wehnert2021legal}, and legal case retrieval \cite{shao2020bert}.

However, when it comes to NER in the legal domain, it remains to be explored in detail, mainly due to the lack of uniform typology of named entities' semantic concepts and the lack of publicly available datasets with named entities annotations \cite{leitner2020dataset}. In 2019, Leitner et al. \cite{leitner2019fine} published their work on this concept that employs CRFs and BiLSTMs and achieved above 90\% F1 scores. Later, they also published the dataset on which they managed to achieve these results. However, there is very minimal research done on using more efficient and improved language models, such as BERT or RoBERTa for the task of NER in the legal domain. In this paper, we use the same dataset to fine-tune a pre-trained German BERT model \cite{chan2020german}. This German BERT model is trained on 6 GB of German Wikipedia, 2.4 GB of OpenLegalData \footnote{\url{https://de.openlegaldata.io/}}, and 3.6 GB of news articles. We make this fine-tuned model publicly available in the HuggingFace library\footnote{\url{https://huggingface.co/}}.

\section{Related work}

Named Entity Recognition and Resolution on legal documents from US courts was performed by \cite{dozier2010named}. These documents consist of US case laws, depositions, pleadings, and other trial documents. The authors used lookup, context rules, and statistical models for the NER task. The lookup method simply creates a list of required named entities and tags all mentioned elements in the list as entities of the given type. This method is susceptible to false negatives due to a lack of contextual cues in lookup taggers. The contextual rules method takes into account the contextual cues, for example, in the legal context, a word sequence followed by ``§" represents a law reference. However, this method requires a large dataset with manual annotations. Statistical models assign weights to cues based on their probabilities and statistical concepts. However, as with the contextual rules method, this also requires a large amount of manually annotated data. The authors developed taggers for Jurisdiction, Court, Title, Doctype, and Judge with F1 of 91.72, 84.70, 81.95, 82.42, and 83.01, respectively.

For German legal documents, legal NER work was performed by \cite{leitner2019fine}. For this purpose, the authors created and published their own open-source dataset consisting of 67,000 sentences and 54,000 annotated entities. The authors used this dataset to train Conditional Random Fields (CRFs) and bidirectional Long-Short Term Memory Networks (BiLSTMs). Experimental results showed that BiLSTMs achieve an F1 score of 95.46 and 95.95 for the fine-grained and coarse-grained classes, respectively. In the same year, \cite{luz2018lener} published their work on Named Entity Recognition in Brazilian Legal Text using LSTM-CRF on the LeNER-Br dataset and reported an F1 score of 97.04 and 88.82 for legislation and legal case entities, respectively. The authors created the LeNER-Br dataset by collecting a total of 66 legal documents from Brazilian courts. To have a baseline performance, the authors first performed experiments on the Paramopama corpus \cite{junior2015paramopama}. 

Based on the success of LSTM-CRF models, many researchers conducted experiments in different languages and reported state-of-the-art performance in their respective languages \cite{paislegalnero,ccetindaug2022named}. As can be seen in these works, a lot of research focused on NER tasks in the legal domain is done by BiLSTMs and CRFs. However, there is only a handful of research is done using transformer-based language models. The following research shows that in most cases these transformer-based language models outperform LSTM-CRF models.

The impact of intradomain fine-tuning of deep language models, namely ELMo \cite{sarzynska2021detecting} and BERT, for Legal NER in Portuguese was studied by \cite{bonifacio2020study}. The authors evaluated language models on three different NER tasks, HAREM \cite{freitas2010second}, LeNER-Br \cite{luz2018lener}, and DrugSeizures-Br\footnote{The public agency for law enforcement and prosecution of crimes in the Brazilian state of Mato Grosso do Sul.}. As for the methodology of their experiments, the authors fine-tuned deep LMs pretrained on general-domain corpus on a legal-domain corpus, and supervised training was done on a NER task. The baseline for these experiments is achieved by skipping the fine-tuning process. Based on the experimental results, the authors conclude that legal-domain language models outperform general-domain language models in the case of LeNER-Br and DruSeizures-Br. It reduces the performance in the case of HAREM.

BERT-BiLSTM-CRF model, proposed by \cite{gu2020named}, first used a pre-trained BERT model to generate word vectors and then fed these vectors to a BiLSTM-CRF model for training. The dataset used by the authors consisted of over 2 million words with legal context. As stated by the authors, this data is collected from the People's Procuratorate case information disclosure network, the judgment document network, the Supreme People's Court trial business guidance cases, the public case published by the Supreme People's Court Gazette, and the judicial dictionary ``Compilation of China's Current Law". The experiment results show their model outperforms BiLSTM, BiLSTM-CFR, and Radical-BILSTM-CRF with 88.86, 87.49, and 87.97 precision, recall, and F1-score, respectively.

\cite{aibek2020named} developed a prototype of the ``Smart Judge Assistant", SJA, recommender system. While developing this prototype, the authors faced the challenge of hiding the personal data of concerned parties. To solve this problem, they used several NER models, namely CRF, LSTM with character embeddings, LSTM-CRF, and BERT, to extract personal information in Russian and Kazakh languages. Out of all five NER models used, BERT shows the highest F1 score of 87.


In addition to the above-mentioned research works, there also exist works in different languages \cite{souza2019portuguese,zanuz2022fostering}. However, to the best of our knowledge, there is yet to be any work done when it comes to developing a transformer-based language model, BERT, for the legal domain in the German language. Therefore, as stated in section \ref{sec:introduction}, in this paper, we aim to use the dataset from \cite{leitner2020dataset} to fine-tune a pre-trained German BERT model.

\section{Dataset}
\label{sec:dataset}

We use the Legal Entity Recognition (LER) dataset published by Leitner et al. in 2020. This dataset was constructed using texts gathered from the XML documents of 750 court decisions from 2017 and 2018 from ``Rechtsprechung im Internet"\footnote{\url{http://www.rechtsprechung-im-internet.de/}}. It includes 107 documents from the following seven federal courts: Federal Labour Court (\textbf{BAG}), Federal Fiscal Court (\textbf{BFH}), Federal Court of Justice (\textbf{BGH}), Federal Patent Court (\textbf{BPatG}), Federal Social Court (\textbf{BSG}), Federal Constitutional Court (\textbf{BVerfG}), and Federal Administrative Court (\textbf{BVerwG}). This data was collected from \textit{Mitwirkung}, \textit{Titelzeile}, \textit{Leitsatz}, \textit{Tenor}, \textit{Tatbestand}, \textit{Entscheidungsgründe}, \textit{Gründen}, \textit{abweichende Meinung}, and \textit{sonstiger Titel} XML elements of corresponding XML documents. As shown in Table \ref{tab:dataset}, it contains a total of 66,723 sentences with 2,157,048 tokens, including punctuation.

\begin{table}[h]
\centering
\begin{tabular}{lcrr}
\hline
\textbf{Court} & \textbf{Documents} & \textbf{Tokens} & \textbf{Sentences}\\
\hline
BAG & 107 & 343,065 & 12,791 \\
BFH & 107 & 276,233 & 8,522 \\
BGH & 108 & 177,835 & 5,858 \\
BPatG & 107 & 404,041 & 12,016 \\
BSG & 107 & 302,161 & 8,083 \\
BVerfG & 107 & 305,889 & 9,237 \\
BVerwG & 107 & 347,824 & 10,216 \\
\hline
\textbf{Total} & \textbf{750} & \textbf{2,157,048} & \textbf{66,723} \\
\hline
\end{tabular}
\caption{Dataset statistics \cite{leitner2020dataset}}
\label{tab:dataset}
\end{table}

This dataset comprises seven coarse-grained classes: \textit{Person} (\textbf{PER}), \textit{Location} (\textbf{LOC}), \textit{Organization} (\textbf{ORG}), \textit{Legal norm} (\textbf{NRM}), \textit{Case-by-case regulation} (\textbf{REG}), \textit{Court decision} (\textbf{RS}), and \textit{Legal literature} (\textbf{LIT}).

These seven coarse-grained classes are then further categorized into 19 fine-grained classes: \textit{Person} (\textbf{PER}), \textit{Judge} (\textbf{RR}), \textit{Lawyer} (\textbf{AN}), \textit{Country} (\textbf{LD}), \textit{City} (\textbf{ST}), \textit{Street} (\textbf{STR}), \textit{Landscape} (\textbf{LDS}), \textit{Organization} (\textbf{ORG}), \textit{Company} (\textbf{UN}), \textit{Institution} (\textbf{INN}), \textit{Court} (\textbf{GRT}), \textit{Brand} (\textbf{MRK}), \textit{Law} (\textbf{GS}), \textit{Ordinance} (\textbf{VO}), \textit{European legal norm} (\textbf{EUN}), \textit{Regulation} (\textbf{VS}), \textit{Contract} (\textbf{VT}), \textit{Court decision} (\textbf{RS}), \textit{Legal literature} (\textbf{LIT}).

Table 1 in \cite{leitner2020dataset} shows the distribution of both classes in the dataset. As shown in that table, classes related to the legal domain, namely \textit{Legal norm}, \textit{Case-by-case regulation}, \textit{Court decision}, and \textit{Legal literature} make up a total of 39,872 annotated NEs, which 74.34\% of the total annotated NEs.

This dataset is publicly available\footnote{\url{https://github.com/elenanereiss/Legal-Entity-Recognition/tree/master/data}} in CoNLL-2002 format \cite{sang2003introduction}. It follows IOB-tagging, where prefix \textit{B-} denotes the beginning of the chunk, prefix \textit{I-} denotes the inside of the chunk, and prefix \textit{O-} denotes the outside of the chunk. In the legal context, consider Table \ref{tab:example}:

\begin{table}[h]
\centering
\begin{tabular}{lc}
\hline
\textbf{Chunk} & \textbf{IOB-Tag}\\
\hline
Das & O \\
Bundesarbeitsgericht & B-GRT \\
ist & O \\
gemäß & O \\
§ & B-GS \\
9Abs. & I-GS \\
2Satz & I-GS \\
2ArbGG & I-GS \\
iVm. & O \\
\hline
\end{tabular}
\caption{An example of IOB-tagging in the legal context. Here, GRT stands for \textit{Court} and GS stands for \textit{Law}.}
\label{tab:example}
\end{table}

\section{Experiment}

\subsection{German BERT}

\begin{figure}[h]
	\centering
	\includegraphics[width=1.0\linewidth]{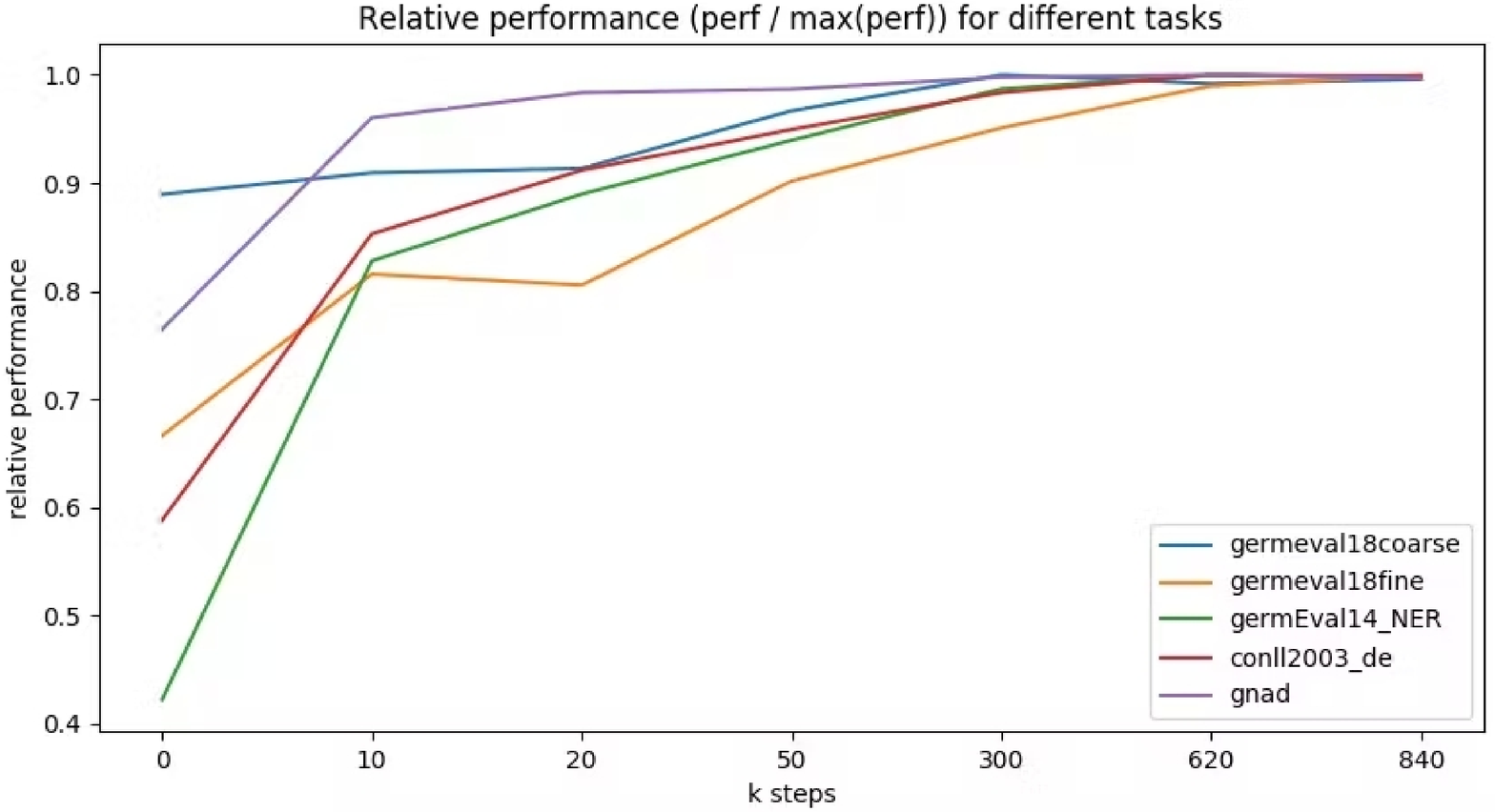}
	\caption{Relative performance of 5 different downstream tasks on 7 models checkpoints\cite{gbert}.}
	\label{fig:rnn_tanh_bm}
\end{figure}

The German BERT model was published by deepset\footnote{\url{https://www.deepset.ai/}} in 2019. As mentioned by the authors of this state-of-the-art BERT model for the German language, it ``\textit{significantly outperforms Google's multilingual BERT model on all 5 downstream NLP tasks we've evaluated}", namely, \textit{germEval18Fine}\footnote{\url{https://github.com/uds-lsv/GermEval-2018-Data}}, \textit{germEval18Coarse}, \textit{germEval14}\footnote{\url{https://sites.google.com/site/germeval2014ner/data}}, \textit{CONLL03}\footnote{\url{https://github.com/MaviccPRP/ger_ner_evals/tree/master/corpora/conll2003}}, and \textit{10kGNAD}\footnote{\url{https://github.com/tblock/10kGNAD}}. Figure \ref{fig:rnn_tanh_bm} shows the relative performance of all five downstream tasks on seven different model checkpoints for up to 840k training steps.

\subsection{Fine-tuning and results}

We compare the performance of our model with the BiLSTM-CRF+ model from \cite{leitner2019fine}.  To ensure that our model is generalized, i.e.that it does not rely on a portion of a dataset, we performed a stratified 10-fold cross-validation. During each cross-validation loop, we use one fold of the dataset as a validation set, while the remaining nine are used for training purposes. In each loop, we fine-tune the German BERT model for seven epochs. This cross-validation also helps in confirming that our model is not over-fitting. 

Table \ref{tab:fine-output} compares the individual performance scores of our model and BiLSTM-CRF+ model for each fine-grained class in the dataset. The reason for choosing the BiLSTM-CRF+ model to compare our performance to is because it has been proven to achieve better performance than CRFs for NER tasks on German legal documents \cite{leitner2019fine}. The higher performance of our fine-tuned model can also be attributed to the fact that one of the datasets used for training the underlying German BERT model comes from OpenLegalData.

\begin{table*}
\centering
\begin{tabular}{l|ccc|ccc}
\hline
\textbf{Class} & \multicolumn{3}{c}{Our model} & \multicolumn{3}{c}{BiLSTM-CRF+} \\
\hline
{} & \textbf{Precision} & \textbf{Recall} & \textbf{F1} & \textbf{Precision} & \textbf{Recall} & \textbf{F1}\\
\hline
Person & 91.48 & 91.09 & 91.29 & 90.78 & 92.24 & \textbf{91.45} \\ 
Judge & 98.72 & 99.53 & \textbf{99.12} & 98.37 & 99.21 & 98.78 \\ 
Lawyer & 96.49 & 85.94 & \textbf{90.91} & 86.18 & 90.59 & 87.07 \\
\hline
Country & 92.51 & 94.2 & 93.34 & 96.52 & 96.81 & \textbf{96.66} \\
City & 88.21 & 89.92 & \textbf{89.06} & 82.58 & 89.06 & 85.60 \\ 
Street & 85.57 & 81.37 & \textbf{83.42} & 81.82 & 75.78 & 77.91 \\ 
Landscape & 68.49 & 68.49 & 68.49 & 78.50 & 80.20 & \textbf{78.25} \\ 
\hline
Organization & 89.11 & 92.22 & \textbf{90.64} & 82.70 & 80.18 & 81.28 \\ 
Company & 97.16 & 97.37 & \textbf{97.27} & 90.05 & 88.11 & 89.04 \\ 
Institution & 94.05 & 94.05 & \textbf{94.05} & 89.99 & 92.40 & 91.17 \\ 
Court & 97.3 & 98.02 & 97.66 & 97.72 & 98.24 & \textbf{97.98} \\ 
Brand & 81.86 & 54.57 & 65.49 & 83.04 & 76.25 & \textbf{79.17} \\
\hline 
Law & 99.36 & 99.23 & \textbf{99.29} & 98.34 & 98.51 & 98.42 \\ 
Ordinance & 94.46 & 96.72 & \textbf{95.58} & 92.29 & 92.96 & 92.58 \\ 
European legal norm & 95.36 & 98.13 & \textbf{96.73} & 92.16 & 92.63 & 92.37 \\
\hline 
Regulation & 89.94 & 87.99 & \textbf{88.95} & 85.14 & 78.87 & 81.63 \\ 
Contract & 96.52 & 95.08 & \textbf{95.79} & 92.00 & 92.64 & 92.31 \\ 
\hline
Court decision & 99.25 & 99.52 & \textbf{99.39} & 96.70 & 96.73 & 96.71 \\
\hline 
Legal literature & 96.91 & 95.57 & \textbf{96.24} & 94.34 & 93.94 & 94.14\\ 
\hline
\end{tabular}
\caption{The performance of our fine-tuned German BERT model and the BiLSTM-CRF+ model for each individual fine-grained class.}
\label{tab:fine-output}
\end{table*}

\subsection{Published fine-tuned model}

\begin{figure}[!h]
	\centering
	\includegraphics[width=1.0\linewidth]{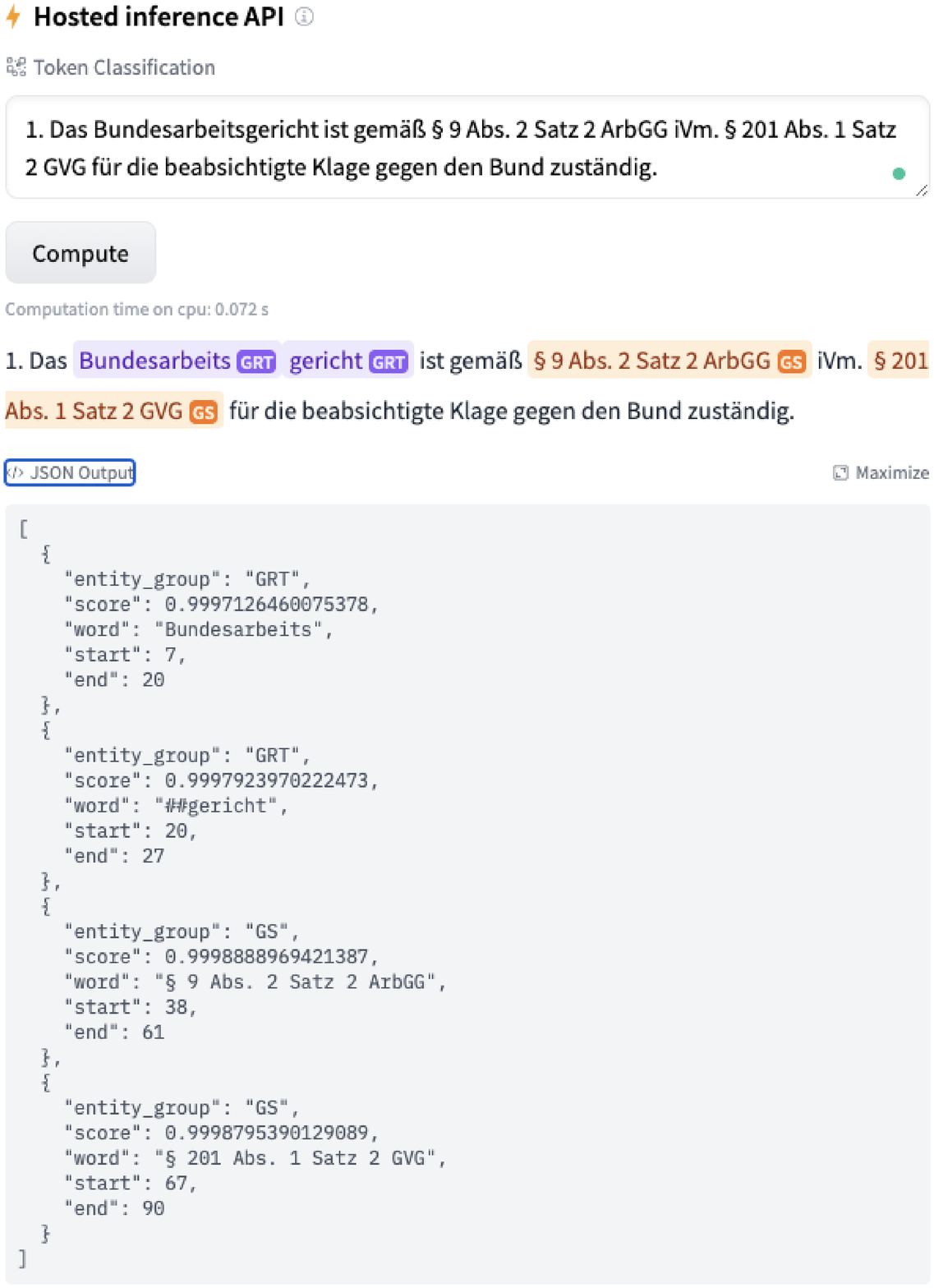}
	\caption{An example output of our German BERT for Legal NER model.}
	\label{fig:gbert-ner}
\end{figure}

Due to the satisfying performance of our fine-tuned German BERT model, we decided to open-source it on HuggingFace\footnote{\url{https://huggingface.co/harshildarji/gbert-legal-ner}}. This model can be used with both popular frameworks, i.e. PyTorch\footnote{\url{https://pytorch.org/}} and TensorFlow\footnote{\url{https://www.tensorflow.org/}}. HuggingFace also provides a ``Hosted inference API"\footnote{\url{https://huggingface.co/docs/api-inference/index}} that allows users to load and test a model in the browser. Figure \ref{fig:gbert-ner} shows an example output of our model via this hosted interface API service.

\section{\uppercase{Conclusions}}
\label{sec:conclusion}

In order to fill the gap of having a proper Legal NER language model in the German language, we fine-tuned a state-of-the-art German BERT on the Legal Entity Recognition dataset on an Nvidia GeForce RTX GPU with a batch size of 64. It took 7 epochs for the fine-tuned model to achieve a very good performance.

If we look at the performance of individual fine-grained entities, in most cases, it outperforms the BiLSTM-CRF+ model used by the authors of the LER dataset. The only classes where our model significantly lags behind are \textit{Country}, \textit{Brand}, and \textit{Landscape}. The performance on these classes can further be improved by having more examples of such instances in the dataset, as currently, only a couple of hundred of them exist compared to \textit{Law} or \textit{Court decisions}.

\vfill
\section*{\uppercase{Acknowledgements}}

\begin{figure}[h]
\includegraphics[width=2.5cm]{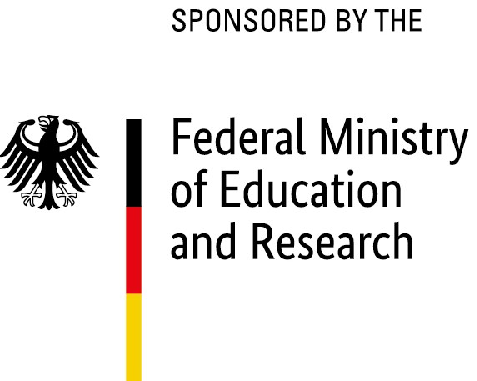}
\end{figure} 

The project on which this report is based was funded by the German Federal Ministry of Education and Research (BMBF) under the funding code 01|S20049, and also partially by the project DEEP WRITE (Grant No. 16DHBKI059). The author is responsible for the content of this publication.

\bibliographystyle{apalike}
{\small
\bibliography{example}}

\end{document}